\documentclass[twocolumn]{svjour3}          % twocolumn
\smartqed  % flush right qed marks, e.g. at end of proof
\usepackage{graphicx, color}

\usepackage{times}
\usepackage{url}
\usepackage{amsmath}
\usepackage{amssymb}
\usepackage{amsfonts,latexsym}
\usepackage{array}
\usepackage{comment}
\usepackage{booktabs} % For formal tables
\usepackage{balance}
\usepackage{hyperref}
\usepackage{tabularx}
\usepackage{cite}

\usepackage{etoolbox}

\makeatletter
\newcommand\urlfootnote@[1]{\footnote{\url@{#1}}}
\DeclareRobustCommand{\urlfootnote}{\hyper@normalise\urlfootnote@}
\makeatother

\usepackage{booktabs, cellspace}

\def\bm#1{\mbox{\boldmath $#1$}}

\journalname{Digital Watermarking for Deep Neural Networks}

\title{Digital Watermarking for Deep Neural Networks}

\author{
Yuki Nagai
\and
Yusuke Uchida 
\and 
Shigeyuki Sakazawa
\and 
Shin'ichi Satoh
}

\institute{Y. Nagai \at
              KDDI Research, Inc. \\
              2-1-15 Ohara, Fujimino-shi, Saitama, 356-8502, Japan\\
              \email{yk-nagai@kddi-research.jp}           %  \\
           \and
           Y. Uchida \at
           DeNA Co., Ltd.\\
           (This work was done when the author was at KDDI Research, Inc.)  \\
           Shibuya Hikarie, 2-21-1 Shibuya, Shibuya-ku, Tokyo, 150-8510, Japan \\
           \email{yusuke.a.uchida@dena.com} 
           \and 
           S. Sakazawa \at
           Osaka Institute of Technology \\
           KDDI Research, Inc. \\
           1-79-1 Kitayama, Hirakata-city, Osaka, 573-0196, Japan \\
           \email{shigeyuki.sakazawa@oit.ac.jp} 
           \and 
           S. Satoh \at
           National Institute of Informatics \\
           2-1-2 Hitotsubashi, Chiyoda-ku, Tokyo, 101-8430, Japan \\
           \email{satoh@nii.ac.jp} 
}

\date{Received: date / Accepted: date}
\begin{document}
\maketitle
\begin{abstract}
Although deep neural networks have made tremendous
progress in the area of multimedia representation, 
training neural models requires a large amount of data and time. 
It is well-known that utilizing trained models 
as initial weights often achieves lower training error than neural networks that are not pre-trained.  
A fine-tuning step helps to reduce both the computational cost and improve performance.
Therefore, sharing trained models has been very important for the rapid progress of research and development.
In addition, trained models could be important assets for the owner(s) who trained them, hence we regard trained models as intellectual property.
In this paper, we propose a digital watermarking technology for ownership authorization of deep neural networks.
First, we formulate a new problem: embedding watermarks into deep neural networks. 
We also define requirements, embedding situations, and attack types on watermarking in deep neural networks.
Second, we propose a general framework for embedding a watermark in model parameters, using a parameter regularizer.
Our approach does not impair the performance of networks into which a watermark is placed because the watermark is embedded while training the host network.
Finally, we perform comprehensive experiments to reveal the potential of watermarking deep neural networks as the basis of this new research effort.
We show that our framework can embed a watermark during the training of a deep neural network from scratch, and during fine-tuning and distilling, without impairing its performance.
The embedded watermark does not disappear even after fine-tuning or parameter pruning; the watermark remains complete even after 65\% of parameters are pruned.
\end{abstract}

\section{Introduction}
Deep neural networks have made tremendous progress in the area of multimedia representation~\cite{wan_mm14, vanden_nips13, feet_eccv14, pang_tom15}. It attempts to model high-level abstractions in data by employing deep architectures composed of multiple non-linear transformations~\cite{bengio_pami13}. In addition, deep neural networks can be applied to various types of data such as sound~\cite{vanden_nips13}, video~\cite{karpathy_eccv14}, text~\cite{mikolov_10}, time series~\cite{zhang_03}, and images~\cite{kri_nips12}. 
In particular, deep convolutional neural networks (DCNN) 
such as LeNet~\cite{lec_ieee98}, \\
AlexNet~\cite{kri_nips12}, VGGNet~\cite{Simonyan_iclr15}, GoogLeNet~\cite{Szegedy_cvpr15}, and ResNet~\cite{He_cvpr16} have demonstrated remarkable performance for a wide range of computer vision problems and other applications.

Additionally, many deep learning frameworks have been released. They help engineers and researchers to develop systems based on deep learning or do research with less effort.
Examples of these great deep learning frameworks are
Caffe~\cite{jia_mm14},
Theano~\cite{bergstra_scipy10},
Torch~\cite{collobert_nipsw11},
Chainer~\cite{tokui_nipsw15},
TensorFlow~\cite{abadi_arxiv16},
and Keras~\cite{chollet_github15}.

Although these frameworks have made it easy to utilize deep neural networks in real applications, training is still a difficult task because it requires a large amount of data and time; for example, several weeks are needed to train a very deep ResNet with the latest GPUs on the ImageNet dataset for instance~\cite{He_cvpr16}.

Therefore, trained models are sometimes provided on web sites in order to make it easy to try out a certain model or reproduce the results in research articles without training.
For example, 
Model Zoo\urlfootnote{https://github.com/BVLC/caffe/wiki/Model-Zoo}
provides trained Caffe models for various tasks with useful utility tools.

It has been empirically observed that utilizing trained models to initialize the weights of a deep neural network has potential the following benefits.
Fine-tuning~\cite{Simonyan_iclr15} is a strategy to directly adapt such already trained models to another application with minimum re-training time.
It was reported that pre-training neural networks often achieves lower training error than neural networks that are not pre-trained~\cite{erhan_jmlr10, hinton_science06}.

Thus, sharing trained models is very important for the rapid progress of research and development of deep neural network systems. In the future, more systematic model-sharing platforms may appear, by analogy with video sharing sites. 
Some digital distribution platforms for purchase and sale of the trained models or even artificial intelligence skills (e.g. Alexa Skills\footnote{\url{https://www.amazon.com/skills/}}) may appear, similar to Google Play or App Store.

In that sense, trained models could be important assets for the owner(s) who trained them.
Dataset quality and quantity directly affect the accuracy of tasks with large networks.
The success of deep neural networks has been achieved not only by algorithms but also through massive amounts of data and computational power. 
Even if the same architecture is employed for different applications, their model weights and  their performance are not be guaranteed to be equal.
For instance, if two applications employ the same architecture such as  AlexNet~\cite{kri_nips12}, and they are trained in the same manner but with a different dataset, the performance would depend on the quality and quantity of the dataset.
Furthermore, a large cost is incurred to create a dataset of sufficient size for specific and realistic tasks.
From the viewpoint of applications, it could be argued that model weights rather than architectures constitute competitive advantage.

We argue that trained models could be treated as \textit{intellectual property}, and 
we believe that providing copyright protection for trained models is a worthwhile challenge.
Discussion on whether or not the copyright law can protect computationally trained models is outside the scope of this paper.
We focus on how to technically protect the copyrights of trained models. 

To this end, we employ a digital watermarking idea, which is used to identify ownership of the copyright of digital content such as images, audio, and videos.
In this paper, we propose a digital watermarking technology for neural networks.
In particular, we propose a general framework to embed a watermark in deep neural networks models to protect intellectual property and detect intellectual property infringement of trained models.
This paper is an extended version of~\cite{uchida_icmr17} with further analysis of attacks on the
watermark. 

\section{Problem Formulation}
Given a model network with or without trained parameters, we define the task of watermark embedding as embedding $T$-bit vector $\bm{b} \in \{0, 1\}^{T}$ into the parameters of one or more layers of the neural network.
We refer to a neural network in which a watermark is embedded as a \textit{host network}, and refer to the task that the host network is originally trying to perform as the \textit{original task}.

In the following, we formulate (1) requirements for an embedded watermark or an embedding method, (2) embedding situations, and (3) expected types of attacks against which embedded watermarks should be robust.

\begin{table*}[tb]
	\centering
	\caption{Requirements for an effective watermarking algorithm in the image and neural network domains.}
	\label{tab:requirement}
	\begin{tabular}{c||p{22em}|p{22em}} \hline
& \multicolumn{1}{c|}{Image domain}	& \multicolumn{1}{c}{Neural networks domain}	\\ \hline
Fidelity	& The quality of the host image should not be degraded by embedding a watermark.	& The effectiveness of the host network should not be degraded by embedding a watermark.	\\ \hline
Robustness	& The embedded watermark should be robust against common signal processing operations such as lossy compression, cropping, resizing, and so on.	& The embedded watermark should be robust against model modifications such as fine-tuning and model compression.	\\ \hline
Capacity	& \multicolumn{2}{l}{An effective watermarking system must have the ability to 
embed a large amount of information.}	\\ \hline
Security	& \multicolumn{2}{l}{A watermark should in general be secret and should not be accessed, read, or modified  by unauthorized parties.}	\\ \hline
Efficiency	& \multicolumn{2}{l}{The watermark embedding and extraction processes should be fast.}	\\ \hline
	\end{tabular} \\
\end{table*}

\subsection{Requirements}
\label{sec:requirements}
Table~\ref{tab:requirement} summarizes the requirements for an effective watermarking algorithm in an image domain~\cite{Hartung_ieee99, Cox08} and a neural network domain.
While both domains share almost the same requirements, \textit{fidelity} and \textit{robustness} are different in image and neural network domains.
For fidelity in an image domain, it is essential to maintain the perceptual quality of the host image while embedding a watermark.
However, in a neural network domain, the parameters themselves are not important.
Instead, the performance of the original task is important.
Therefore, it is essential to maintain the performance of the trained host network, and not to hamper the training of a host network.

Regarding robustness, as images are subject to various signal processing operations, an embedded watermark should stay in the host image even after these operations.
Note that the greatest possible modification to a neural network is fine-tuning or transfer learning~\cite{Simonyan_iclr15}.
An embedded watermark in a neural network should be detectable after fine-tuning or other possible modifications.

\subsection{Embedding Situations}
\label{sec:situation}
We classify the embedding situations into three types: train-to-embed, fine-tune-to-embed, and distill-to-embed, as summarized in Table~\ref{tab:settings}.

\textbf{Train-to-embed} is the case in which the host network is trained from scratch while embedding a watermark where labels for training data are available.

\textbf{Fine-tune-to-embed} is the case in which a watermark is embedded while fine-tuning.
In this case, model parameters are initialized with a pre-trained network.
The network configuration near the output layer may be changed before fine-tuning in order to adapt the final layer's output to another task.

\textbf{Distill-to-embed} is the case in which a watermark is embedded into a trained network \textit{without} labels using the distilling approach~\cite{hin_nipsw14}.
Embedding is performed in fine-tuning where the predictions of the trained model are used as labels.
In the standard distill framework, a large network (or multiple networks) is first trained and then a smaller network is trained using the predicted labels of the large network in order to compress the large network.
In this paper, we use the distill framework as a simple way to train a network without labels.

The first two situations assume that the copyright holder of the host network is expected to embed a watermark into the host network during training or fine-tuning.
Fine-tune-to-embed is also useful when a model owner wants to embed individual watermarks to identify those to whom the model had been distributed.
By doing so, individual instances can be tracked.
The last situation assumes that a non-copyright holder (e.g., a platformer) is entrusted to embed a watermark on behalf of a copyright holder.

\begin{table}[tb]
	\centering
	\caption{Three embedding situations. Fine-tune indicates whether parameters are initialized in embedding using already trained models, or not. Label availability indicates whether or not labels for training data are available in embedding.}
	\label{tab:settings}
	\begin{tabular}{c|cc} \hline
				& Fine-tune		& Label availability	\\ \hline
Train-to-embed		&				& \checkmark	\\
Fine-tune-to-embed	& \checkmark	& \checkmark	\\
Distill-to-embed	& \checkmark	&				\\ \hline
	\end{tabular} \\
\end{table}

\subsection{Expected Attack Types}
\label{sec:attack}
Related to the requirement for robustness in Section~\ref{sec:requirements}, we assume three types of attacks against which embedded watermarks should be robust: fine-tuning, model compression and watermark overwriting.

\subsubsection{Fine-tuning} 
Fine-tuning~\cite{Simonyan_iclr15} seems to be the most feasible type of attack, whether intentionally or unintentionally, because it  empirically has the following potential benefits as follows.
To utilize trained models as initial weights of training another networks often achieves lower training error than training from scratch~\cite{erhan_jmlr10, hinton_science06}.
The fine-tuning step helps to reduce both the computational cost and improve the performance.
Many models have been constructed on top of existing state-of-the-art models.
Fine-tuning alters the model parameters, and thus embedded watermarks should be robust against this alteration.

\subsubsection{Model compression} 
Model compression is very important in deploying deep neural networks in embedded systems or mobile devices as it can significantly reduce memory requirements and/or computational cost.
Model compression can be easily imagined by analogy with lossy image compression in the image domain.
Lossy compression distorts model parameters, so we should explore how it affects the detection rate.

\begin{comment}
\subsubsection{Distillation} 
Distillation as an attack is a very specific to deep neural networks. Distillation is employed for various purposes such as reducing computational complexity or compressing the knowledge in an ensemble of models into a single model. It could be unintentional attack as well as fine-tuning.
\end{comment}

\subsubsection{Watermark overwriting} 
Watermark overwriting would be a severe attack.
Attackers may try to destroy an existing watermark by embedding different watermark in the same manner.
Ideally embedded watermarks should be robust against this type of attack.

\section{Proposed Framework}
In this section, we propose a framework for embedding a watermark into a host network.
Although we focus on a DCNN~\cite{lec_ieee98} as the host, our framework is essentially applicable to other networks such as standard multilayer perceptron (MLP), recurrent neural networks (RNN), and long short-term memory (LSTM)~\cite{hoch_nc1997}.

\subsection{Embedding Targets}
In this paper, a watermark is assumed to be embedded into one of the convolutional layers in a host DCNN\footnote{Fully-connected layers can also be used but we focus on convolutional layers here, because fully-connected layers are often discarded in fine-tuning.}.
Let $(S, S)$, $D$, and $L$ respectively denote the size of the convolution filter, the depth of input to the convolutional layer, and the number of filters in the convolutional layer.
The parameters of this convolutional layer are characterized by the tensor $\bm{W} \in \mathbb{R}^{S \times S \times D \times L}$.
The bias term is ignored here.
Let us think of embedding a $T$-bit vector $\bm{b} \in \{0, 1\}^{T}$ into $\bm{W}$.
The tensor $\bm{W}$ is a set of $L$ convolutional filters and the order of the filters does not affect the output of the network if the parameters of the subsequent layers are appropriately re-ordered.
In order to remove this arbitrariness in the order of filters, we calculate the mean of $W$ over $L$ filters as $\overline{W}_{ijk} = \tfrac{1}{L} \sum_l W_{ijkl}$.
Letting $\bm{w} \in \mathbb{R}^M$ ($M = S \times S \times D$) denote a flattened version of $\overline{\bm{W}}$, our objective is now to embed $T$-bit vector $\bm{b}$ into $\bm{w}$.

\subsection{Embedding Regularizer}
\label{sec:regularizer}
It is possible to embed a watermark into a host network by directly modifying $\bm{w}$ of a trained network, as is usually done in the image domain.
However, this approach degrades the performance of the host network in the original task as shown later in Section~\ref{sec:direct}.
Instead, we propose embedding a watermark while \textit{training} a host network for the original task so that the existence of the watermark does not impair the performance of the host network in its original task.
To this end, we utilize a \textit{parameter regularizer}, which is an additional term in the original cost function for the original task.
The cost function $E(\bm{w})$ with a regularizer is defined as:
\begin{equation}
\label{eq:param}
E(\bm{w}) = E_0 (\bm{w}) + \lambda E_R (\bm{w}),
\end{equation}
where $E_0 (\bm{w})$ is the original cost function, $E_R (\bm{w})$ is a regularization term that imposes a certain restriction on parameters $\bm{w}$, and $\lambda$ is an adjustable parameter.
A regularizer is usually used to prevent overfitting in neural networks.
$L_2$ regularization (or weight decay~\cite{kro_nips92}), $L_1$ regularization, and their combination are often used to reduce over-fitting of parameters for complex neural networks.
For instance, $E_R (\bm{w}) = ||\bm{w}||^2_2$ in the $L_2$ regularization.

In contrast to these standard regularizers, our regularizer imposes a certain statistical bias on parameter $\bm{w}$, as a watermark in a training process.
We refer to this regularizer as an \textit{embedding regularizer}.
Before defining the embedding regularizer, we explain how to extract a watermark from $\bm{w}$.
Given a (mean) parameter vector $\bm{w} \in \mathbb{R}^M$ and an embedding parameter $\bm{X} \in \mathbb{R}^{T{\times}M}$, the watermark extraction is simply done by projecting $\bm{w}$ using $\bm{X}$, followed by thresholding at 0.
More precisely, the $j$-th bit is extracted as:
\begin{equation}
b_j = s( \sum_{i} X_{ji} w_i),
\end{equation}
where $s(x)$ is a step function:
\begin{equation}
	s(x) =
	\begin{cases}
		\, 1 & x \ge 0 \\
		\, 0 & \mathrm{else}.
	\end{cases}
\end{equation}
This process can be considered to be a binary classification problem with a single-layer perceptron (without bias)\footnote{Although this single-layer perceptron can be \textit{deepened} into multi-layer perceptron, we focus on the simplest one in this paper.}.
Therefore, it is straightforward to define the loss function $E_R (\bm{w})$ for the embedding regularizer by using (binary) cross entropy:
\begin{equation}
E_R (\bm{w}) = - \sum_{j=1}^{T} \left( b_j \log(y_j) + (1 - b_j) \log(1 - y_j) \right),
\end{equation}
where $y_j = \sigma(\sum_{i} X_{ji} w_i)$ and $\sigma(\cdot)$ is the sigmoid function:
\begin{equation}
\sigma(x)=\frac{1}{1+\exp(-x)}.
\end{equation}
We call this loss function an \textit{embedding loss} function.

Note that an embedding loss function is used to update $\bm{w}$, not $\bm{X}$, in our framework.
It may be confusing that $\bm{w}$ is an input and $\bm{X}$ is a parameter to be learned in a standard perceptron.
In our case, $\bm{w}$ is an embedding target and $\bm{X}$ is a fixed parameter. $\bm{X}$ works as a secret key~\cite{Hartung_ieee99} to detect an embedded watermark.
The design of $\bm{X}$ is discussed in Section~\ref{sec:param}.

This approach does not impair the performance of the host network in the original task as confirmed in experiments, 
because deep neural networks are typically \\
over-parameterized.
It is well-known that deep neural networks have many local minima, and that all local minima are likely to have an error very close to that of the global minimum~\cite{dauphin_nips14, cho_aistats15}.
Therefore, the embedding regularizer only needs to \textit{guide} model parameters to one of a number of \textit{good} local minima so that the final model parameters have an arbitrary watermark.

\subsection{Regularizer Parameters}
\label{sec:param}
In this section we discuss the design of the embedding parameter $\bm{X}$, which can be considered as a secret key~\cite{Hartung_ieee99} in detecting and embedding watermarks.
While $\bm{X} \in \mathbb{R}^{T{\times}M}$ can be an arbitrary matrix, it will affect the performance of an embedded watermark because it is used in both embedding and extraction of watermarks.
In this paper, we consider three types of $\bm{X}$: $\bm{X}^{\textsf{direct}}$, $\bm{X}^{\textsf{diff}}$, and $\bm{X}^{\textsf{random}}$.

$\bm{X}^{\textsf{direct}}$ is constructed so that one element in each row of $\bm{X}^{\textsf{direct}}$ is '1' and the others are '0'.
In this case, the $j$-th bit $b_j$ is \textit{directly} embedded in a certain parameter $w_{\hat{i}}$ s.t. $\bm{X}^{\textsf{direct}}_{j\hat{i}} = 1$.

$\bm{X}^{\textsf{diff}}$ is created so that each row has one '1' element and one '-1' element, and the others are '0'.
Using $\bm{X}^{\textsf{diff}}$, the $j$-th bit $b_j$ is embedded into the \textit{difference} between $w_{i_+}$ and $w_{i_-}$ where $\bm{X}^{\textsf{diff}}_{ji_+}=1$ and $\bm{X}^{\textsf{diff}}_{ji_-}=-1$.

Each element of $\bm{X}^{\textsf{random}}$ is independently drawn from the standard normal distribution $\mathcal{N}(0, 1)$.
Using $\bm{X}^{\textsf{random}}$, each bit is embedded into all instances of the parameter $w$ with \textit{random} weights.
These three types of embedding parameters are compared in experiments.

\section{Experiments}
In this section, we demonstrate that our embedding regularizer can embed a watermark without impairing the performance of the host network, and the embedded watermark is robust against various types of attacks. Our implementation of the embedding regularizer is publicly available~\footnote{\url{https://github.com/yu4u/dnn-watermark}}.

\subsection{Evaluation Settings}

\subsubsection{Dataset}
For experiments, we used the well-known CIFAR-10 and Caltech-101 datasets.
The CIFAR-10 dataset~\cite{kri_tech09} consists of 60,000 $32 \times 32$ color images in 10 classes, with 6,000 images per class.
These images were separated into 50,000 training images and 10,000 test images.
The Caltech-101 dataset~\cite{fei_gmbv04} includes pictures of objects belonging to 101 categories; it contains about 40 to 800 images per category.
The size of each image is roughly $300 \times 200$ pixels but we resized them to $32 \times 32$ for fine-tuning.
For testing, we used 30 images for training and at most 40 of the remaining images for each category.

\subsubsection{Host Network and Training Settings}
We used the wide residual network~\cite{zag_eccv16} as the host network.
The wide residual network is an efficient variant of the residual network~\cite{He_cvpr16}.
Table~\ref{tab:network} shows the structure of the wide residual network. A depth parameter $N$ is the number of blocks in groups, and a width parameter $k$ is widening
factor that scales the width of the residual blocks in groups.

In all our experiments, we set $N = 1$ and $k = 4$, and used SGD with Nesterov momentum~\cite{amari_67, yurii_83, sutskever_icml13} and cross-entropy loss in training.
The initial learning rate was set at 0.1, weight decay to $5.0{\times}10^{-4}$, momentum to 0.9 and minibatch size to 64.
The learning rate was dropped by a factor of 0.2 at 60, 120 and 160 epochs, and we trained for a total of 200 epochs, following the settings used in~\cite{zag_eccv16}.

We embedded a watermark into one of the following convolution layers: the second convolutional layer in the \textsf{conv 2}, \textsf{conv 3}, and \textsf{conv 4} groups.
Hereinafter, we refer to the location of the host layer by simply describing the \textsf{conv 2}, \textsf{conv 3}, or \textsf{conv 4} group.
In Table~\ref{tab:network}, the number $M$ of parameter $\bm{w}$ is also shown for these layers.
The parameter $\lambda$ in Eq.~(\ref{eq:param}) is set to $0.01$.
As a watermark, we embedded $\bm{b} = \mathbf{1} \in \{0, 1\}^{T}$ in the following experiments.

\begin{comment}
\begin{table}[tb]
	\centering
	\caption{Structure of the host networks. $N$ is the number of blocks in groups and $k$ is a widening
factor that scales the width of blocks.}
	\label{tab:network}
	\begin{tabular}{c|c|c} \hline
Group	& Output size	& Building block	\\ \hline
conv 1	& $32 \times 32$	& $[3 \times 3, 16]$	\\
conv 2	& $32 \times 32$	& $\begin{bmatrix} 3 \times 3, 16 \times k
\\ 3 \times 3, 16 \times k \end{bmatrix} \times N$	\\
conv 3	& $16 \times 16$	& $\begin{bmatrix} 3 \times 3, 32 \times k
\\ 3 \times 3, 32 \times k \end{bmatrix} \times N$	\\
conv 4	& $8 \times 8$	& $\begin{bmatrix} 3 \times 3, 64 \times k
\\ 3 \times 3, 64 \times k \end{bmatrix} \times N$	\\
avg-pool	& $1 \times 1$	& $[8 \times 8]$	\\ \hline
	\end{tabular} \\
\end{table}
\end{comment}

\begin{table}[tb]
	\centering
	\caption{Structure of the host network. $N$ is the number of blocks and $k$ is a widening
factor in groups.}
	\label{tab:network}
	\begin{tabular}{c|c|c|c} \hline
Group	& Output size	& Building block 	& $M$	\\ 
		&			& ResNe block type = $B(3,3)$ & \\ \hline
conv 1	& $32 \times 32$	& $[3 \times 3, 16]$	& N/A	\\
conv 2	& $32 \times 32$	& $\begin{bmatrix} 3 \times 3, 16 \times k
\\ 3 \times 3, 16 \times k \end{bmatrix} \times N$	& $144 \times k$	\\
conv 3	& $16 \times 16$	& $\begin{bmatrix} 3 \times 3, 32 \times k
\\ 3 \times 3, 32 \times k \end{bmatrix} \times N$	& $288 \times k$	\\
conv 4	& $8 \times 8$	& $\begin{bmatrix} 3 \times 3, 64 \times k
\\ 3 \times 3, 64 \times k \end{bmatrix} \times N$	& $576 \times k$	\\
%avg-pool	& $1 \times 1$	& $[8 \times 8]$	& N/A \\ \hline
			& $1 \times 1$	& avg-pool, fc, soft-max	& N/A \\ \hline
	\end{tabular} \\
\end{table}

\subsection{Embedding Results}
We trained the host network from scratch (\textsf{train-to-embed}) on the CIFAR-10 dataset with and without embedding a watermark.
In the embedding case, a 256-bit watermark ($T=256$) was embedded into the \textsf{conv 2} group.

\subsubsection{Detecting Watermarks}
\begin{figure*}[tb]
	\begin{minipage}[b]{0.33\linewidth}
	\centering{\includegraphics[width=\linewidth]{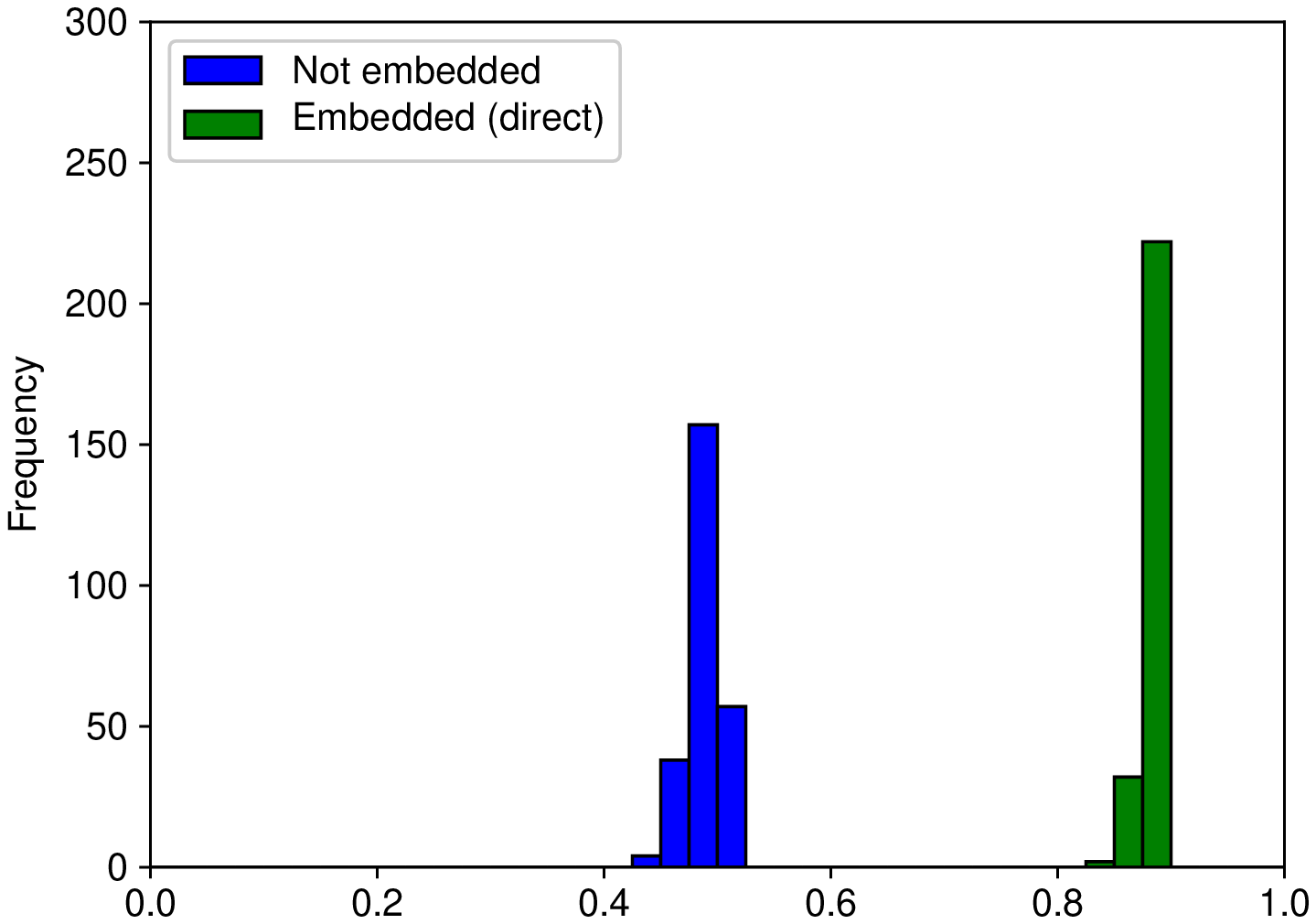} }
		\centerline{(a) \textsf{direct}}
	\end{minipage}
	\begin{minipage}[b]{0.33\linewidth}
	\centering{\includegraphics[width=\linewidth]{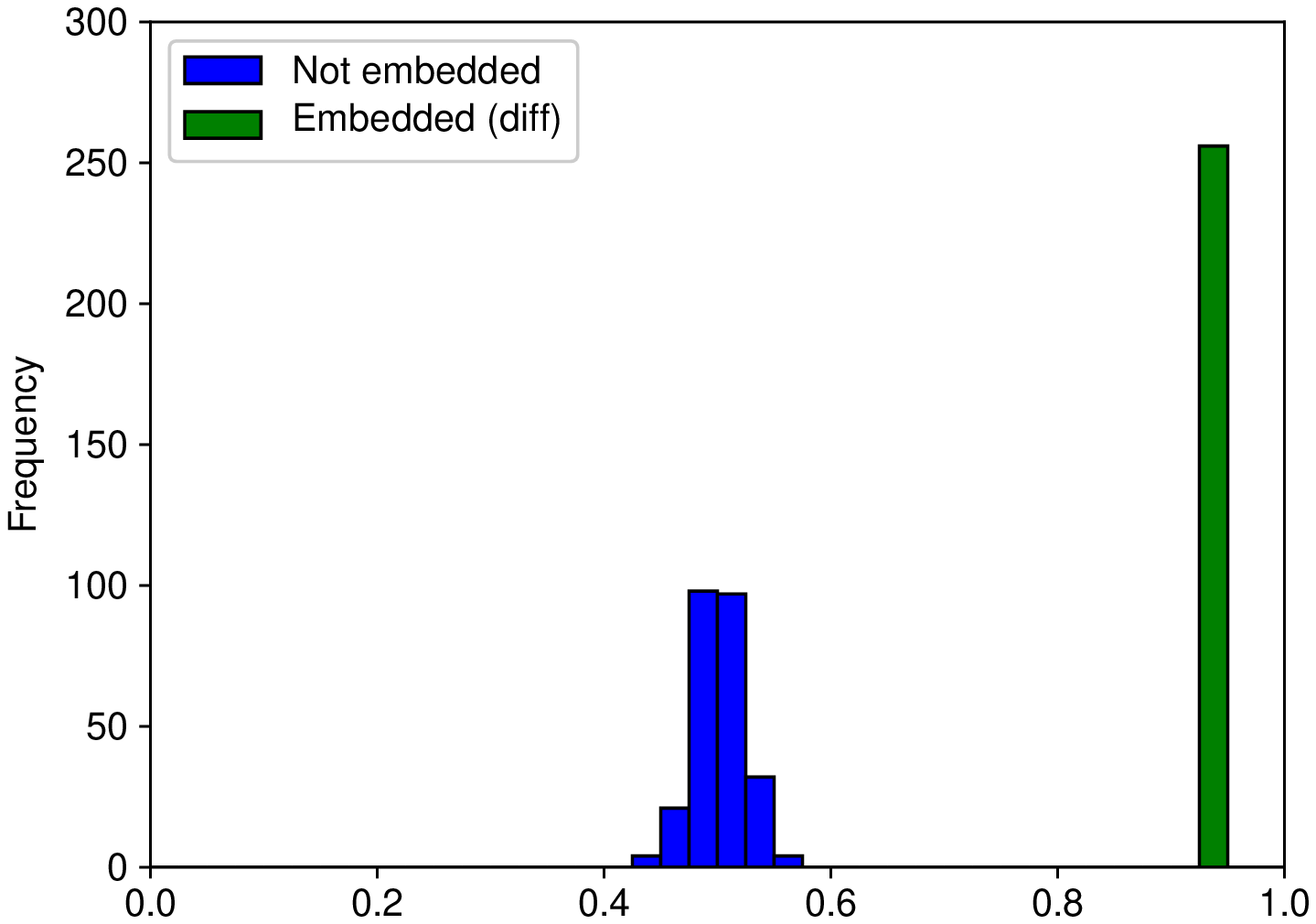}}
	\centerline{(b) \textsf{diff}}
	\end{minipage}  
	\begin{minipage}[b]{0.333\linewidth}
	\centering{\includegraphics[width=\linewidth]{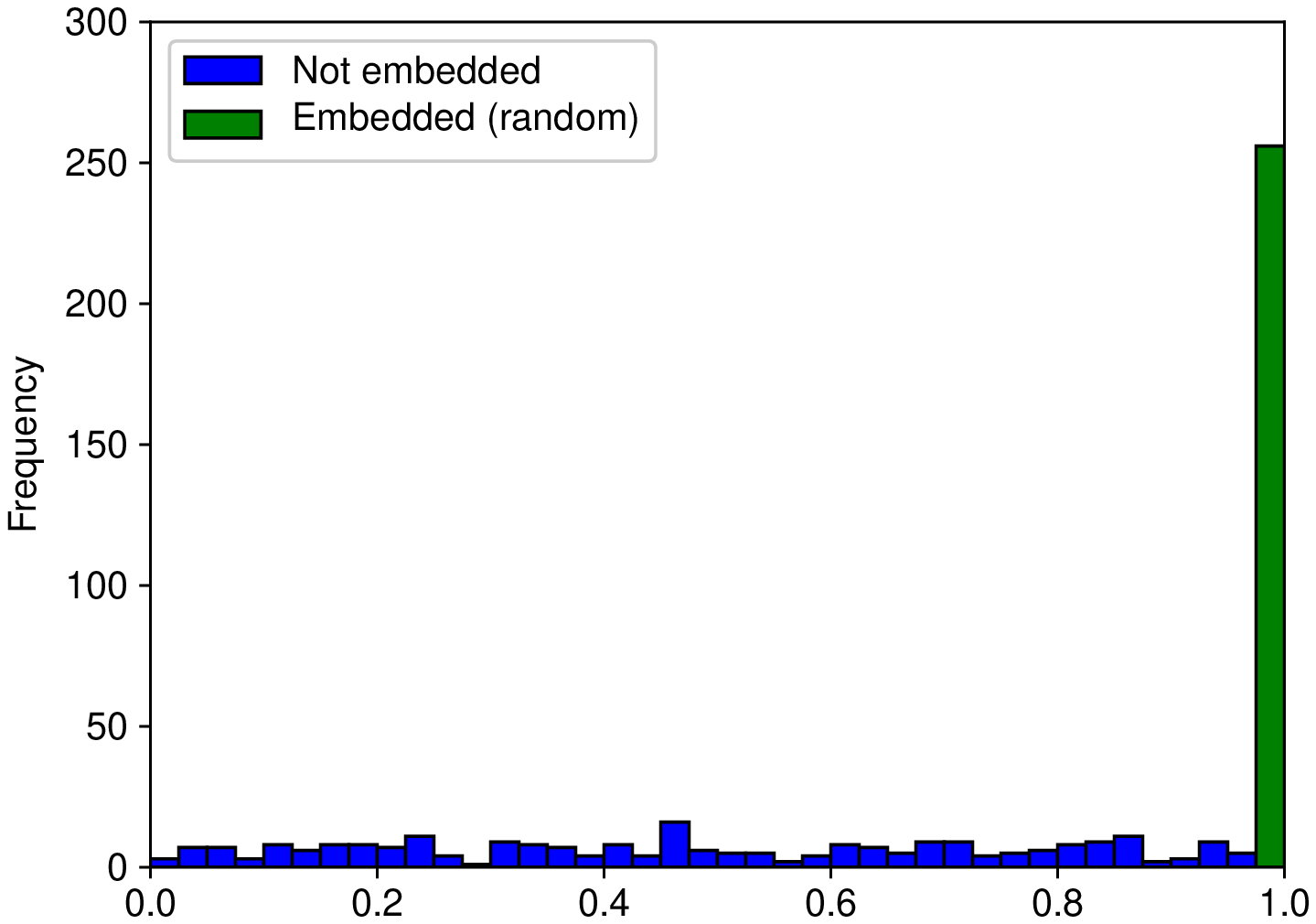} }
	\centering{ (c) \textsf{random}}
	\end{minipage}
	\caption{Histogram of the embedded watermark $\sigma(\sum_{i} X_{ji} w_i)$ (before thresholding) with and without watermarks. All watermarks will be successfully detected by binarizing $\sigma(\sum_{i} X_{ji} w_i)$ at a threshold of 0.5. In the case of \textsf{random}, it can be easily determined whether or not a watermark is embedded with the histogram. }
	\label{fig:hist}
\end{figure*}

Figure~\ref{fig:hist} shows the histogram of the embedded watermark $\sigma(\sum_{i} X_{ji} w_i)$ (before thresholding) with and without watermarks where (a) \textsf{direct}, (b) \textsf{diff}, and (c) \textsf{random} parameters are used in embedding and detection.
If we binarize $\sigma(\sum_{i} X_{ji} w_i)$ at a threshold of 0.5, all watermarks are correctly detected because $\forall j, \; \sigma(\sum_{i} X_{ji} w_i) \ge 0.5$ if and only if $\sum_{i} X_{ji} w_i \ge 0$ for all embedded cases.
Please note that we embedded $\bm{b} = \mathbf{1} \in \{0, 1\}^{T}$ as a watermark as previously mentioned.
Although random watermarks will be detected for the non-embedded cases, it can be easily determined if that the watermark is not embedded because the distribution of $\sigma(\sum_{i} X_{ji} w_i)$ is quite different from those for embedded cases.

\subsubsection{Distribution of Model Parameters}
\begin{figure*}[tb]
	\centering
	\begin{minipage}[b]{0.45\linewidth}
	\includegraphics[width=\linewidth]{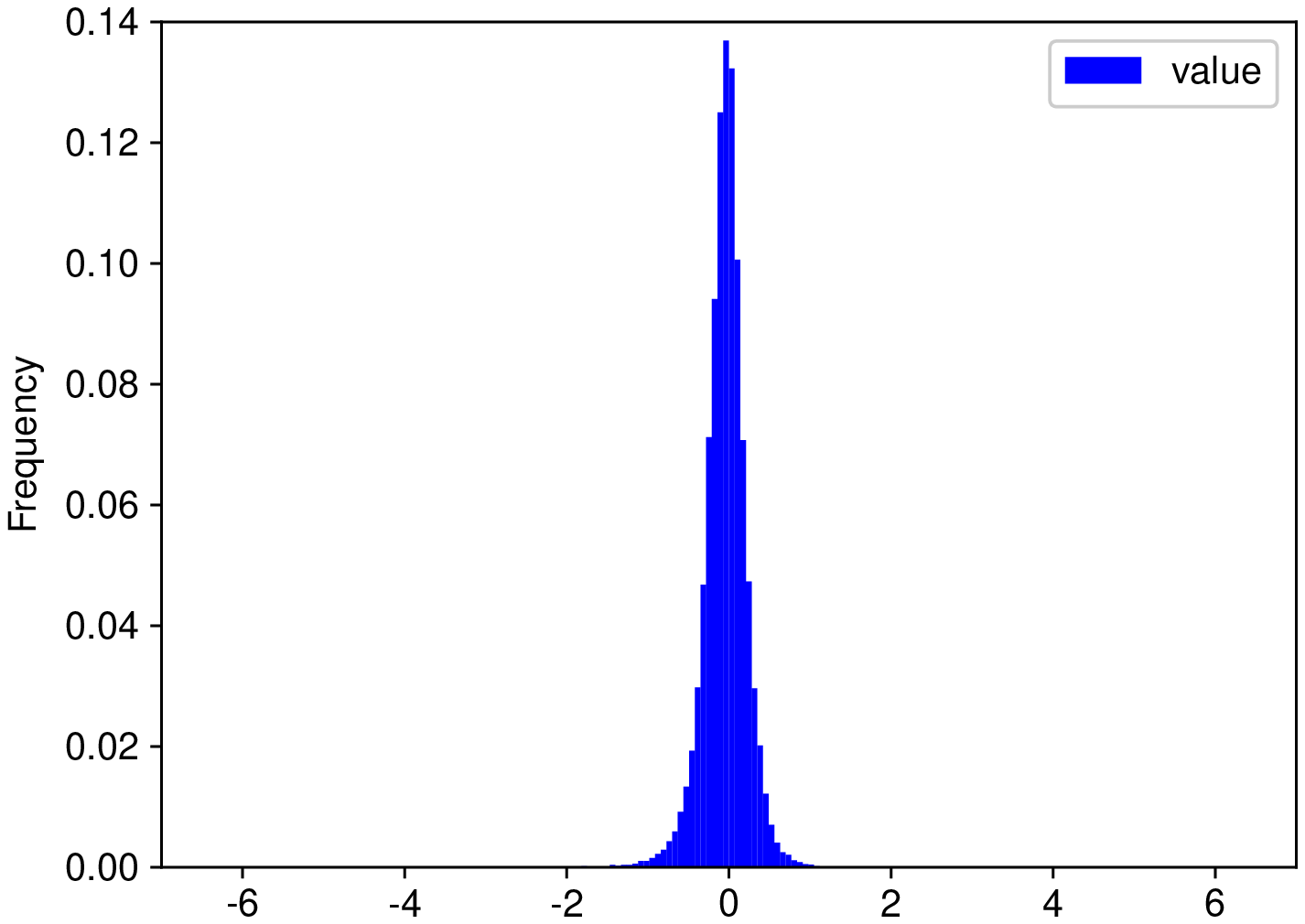} \\ 
	\centering{ (a) \textsf{Not embedded}}
	\end{minipage}
	\begin{minipage}[b]{0.45\linewidth}
		\includegraphics[width=\linewidth]{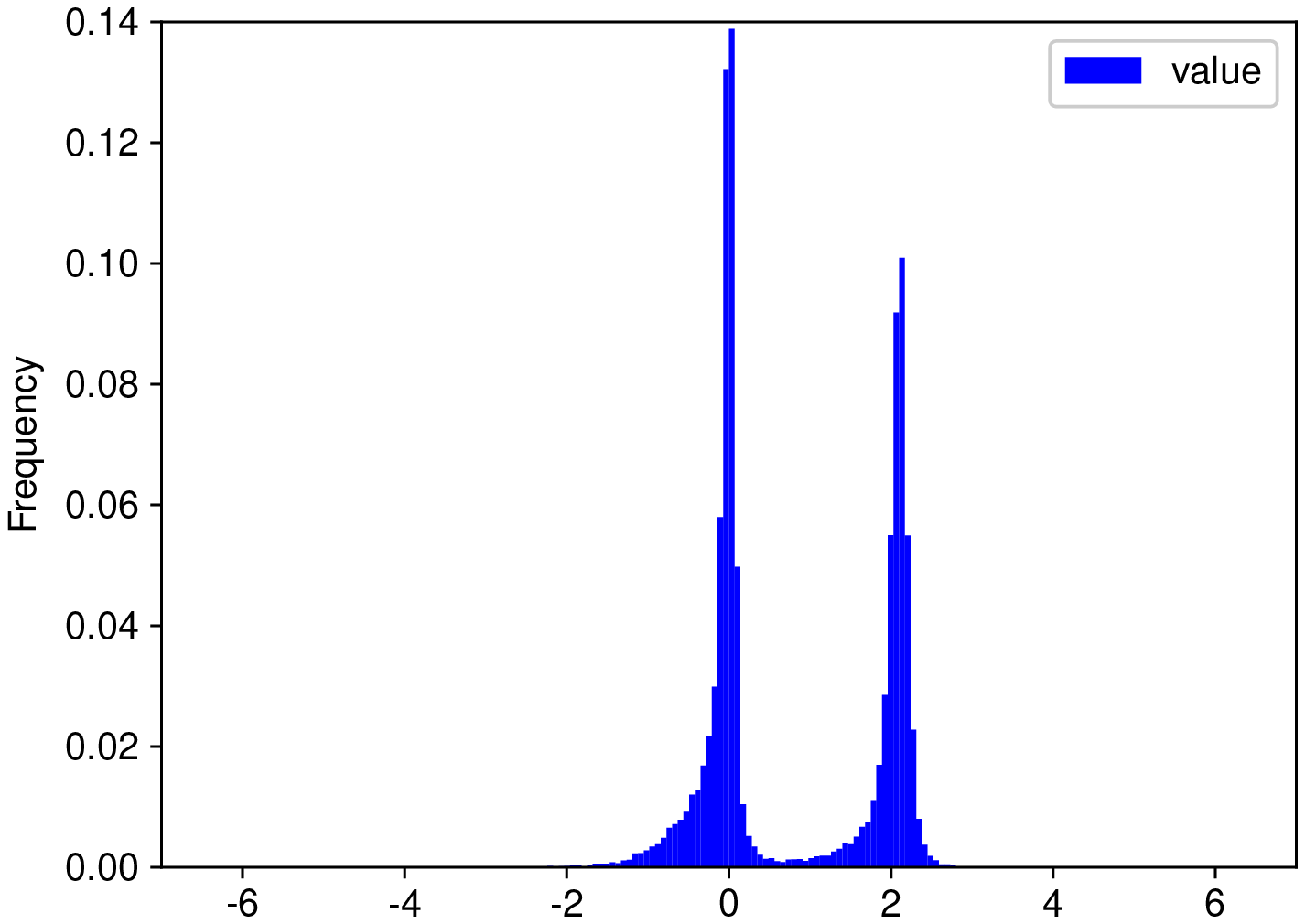} \\
	\centering{ (b) \textsf{direct}}
	\end{minipage} \\
	\begin{minipage}[b]{0.45\linewidth}
		\includegraphics[width=\linewidth]{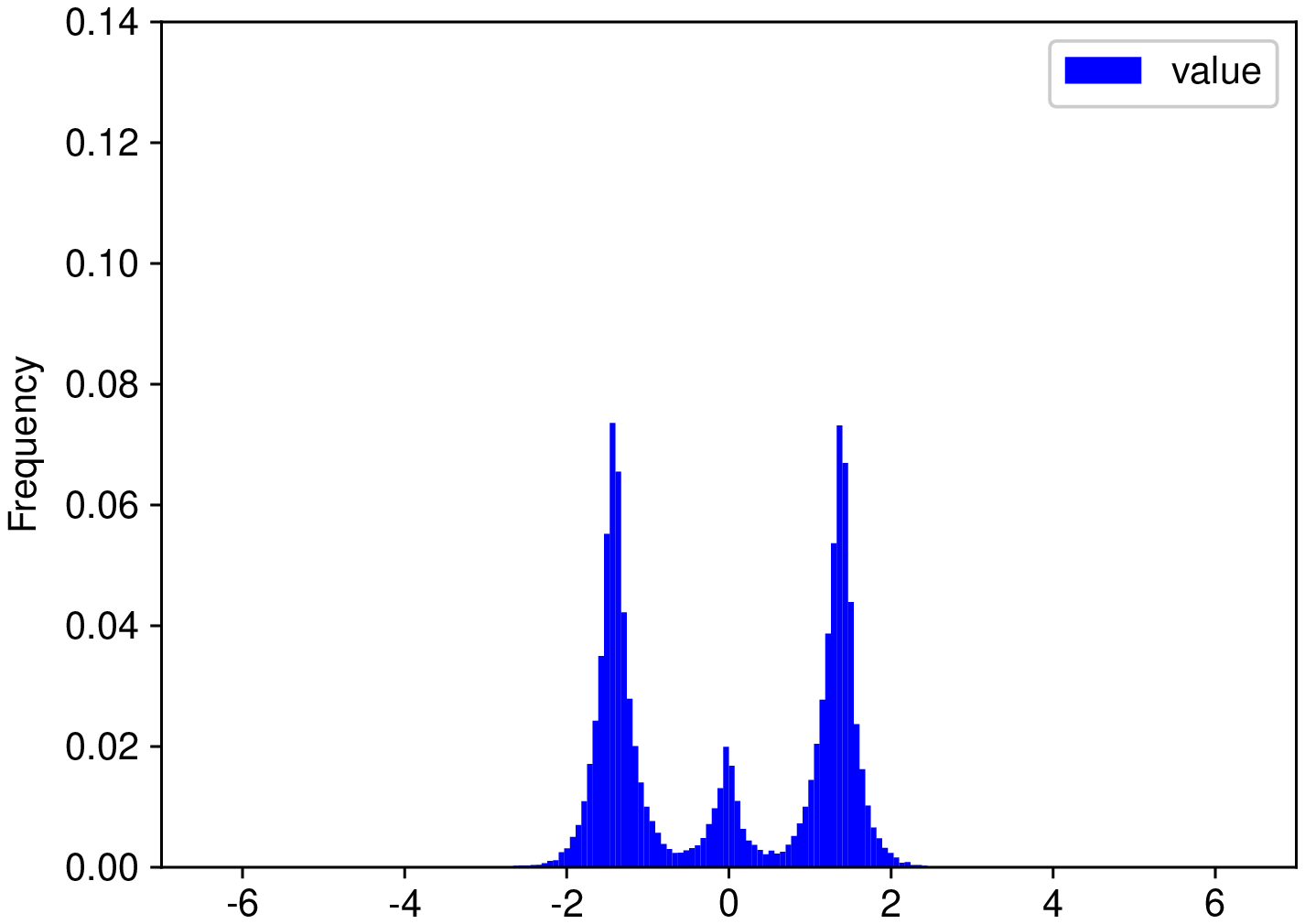} \\
	\centering{ (c) \textsf{diff}}
	\end{minipage}
	\begin{minipage}[b]{0.45\linewidth}
	\includegraphics[width=\linewidth]{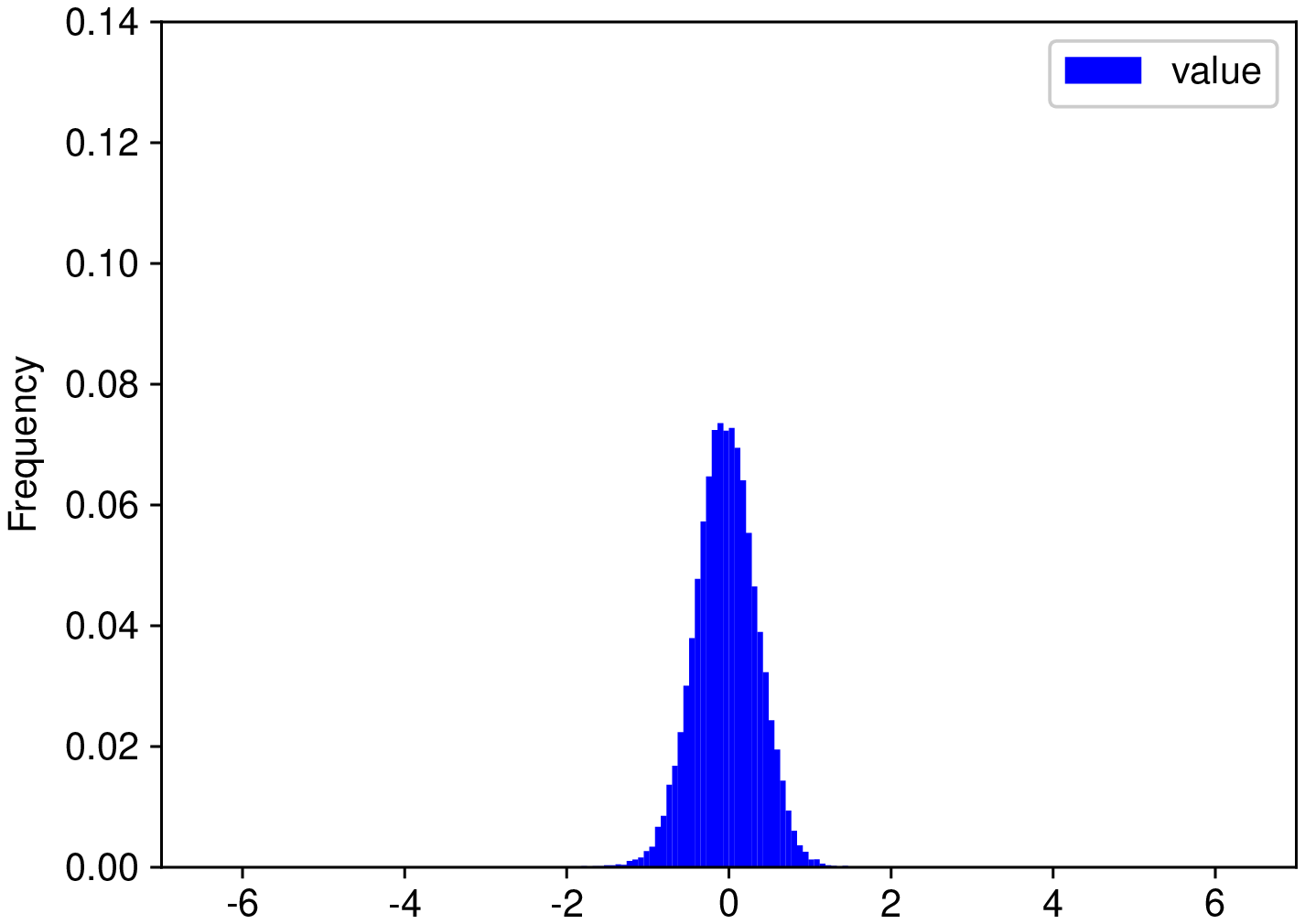} \\
	\centering{ (d) \textsf{random}}
	\end{minipage}
	\caption{Distribution of model parameters $\bm{W}$ with and without watermarks.}
	\label{fig:param_hist}
\end{figure*}

We explore how trained model parameters are affected by the embedded watermarks.
Figure~\ref{fig:param_hist} shows the distribution of model parameters $\bm{W}$ (not $\bm{w}$) with and without watermarks.
These parameters are taken only from the layer in which a watermark was embedded.
Note that $\bm{W}$ is the parameter before taking the mean over filters, and thus the number of parameters is $3 \times 3 \times 64 \times 64$.
We can see that \textsf{direct} and \textsf{diff} significantly alter the distribution of parameters while \textsf{random} does not.
In \textsf{direct}, many parameters became large and a peak appears near 2 so that their mean over filters becomes a large positive value to reduce the embedding loss.
In \textsf{diff}, most parameters were pushed in both positive and negative directions so that the differences between these parameters became large.
In \textsf{random}, a watermark is diffused over all parameters with random weights and thus does not significantly alter the distribution.
This is one of the desirable properties of watermarking related to the security requirement; one may be aware of the existence of the embedded watermarks for the \textsf{direct} and \textsf{diff} cases.

The results so far indicated that the \textsf{random} approach seemed to be the best choice among the three, with low embedding loss, low test error in the original task, and no alteration of  the parameter distribution.
Therefore, in the following experiments, we used the \textsf{random} approach in embedding watermarks without explicitly indicating it.

\subsection{Fidelity}
\label{sec:fidelity}
\subsubsection{Embedding without Training}
\label{sec:direct}
As mentioned in Section~\ref{sec:regularizer}, it is possible to embed a watermark in a host network by directly modifying the trained parameter $\bm{w_0}$ as usually done in the image domain.
Here we try to do this by minimizing the following loss function instead of Eq.~(\ref{eq:param}):
\begin{equation}
\label{eq:direct}
E(\bm{w}) = \tfrac{1}{2} ||\bm{w} - \bm{w}_0||^2_2 + \lambda E_R (\bm{w}),
\end{equation}
where the embedding loss $E_R (\bm{w})$ is minimized while minimizing the difference between the modified parameter $\bm{w}$ and the original parameter $\bm{w}_0$.
Table~\ref{tab:direct_embed} summarizes the embedding results after minimizing Eq.~(\ref{eq:direct}) against the host network trained on the CIFAR-10 dataset.
We can see that embedding fails for $\lambda \le 1$ as the bit error rate (BER) is larger than zero while the test error of the original task becomes too large for  $\lambda > 1$.
Thus, it is not effective to directly embed a watermark without considering the original task.

\begin{table}[tb]
	\centering
	\caption{Losses, test error ($\%$), and bit error rate (BER) after embedding a watermark with different $\lambda$.}
	\label{tab:direct_embed}
\begin{tabular}{c|cccc}  \hline
$\lambda$	& $\tfrac{1}{2} ||\bm{w} - \bm{w}_0||^2_2$	& $E_R (\bm{w})$	& Test error	& BER	\\ \hline
0			& 0.000								& 1.066		& 8.04			& 0.531	\\
1			& 0.184								& 0.609		& 8.52			& 0.324	\\
10			& 1.652								& 0.171		& 10.57			& 0.000	\\
100			& 7.989								& 0.029		& 13.00			& 0.000	\\ \hline
	\end{tabular} \\
\end{table}

\subsubsection{Test Error and Training Loss}
\label{sec:res_train}
\begin{figure}[tb]
	\centering
	\includegraphics[width=\linewidth]{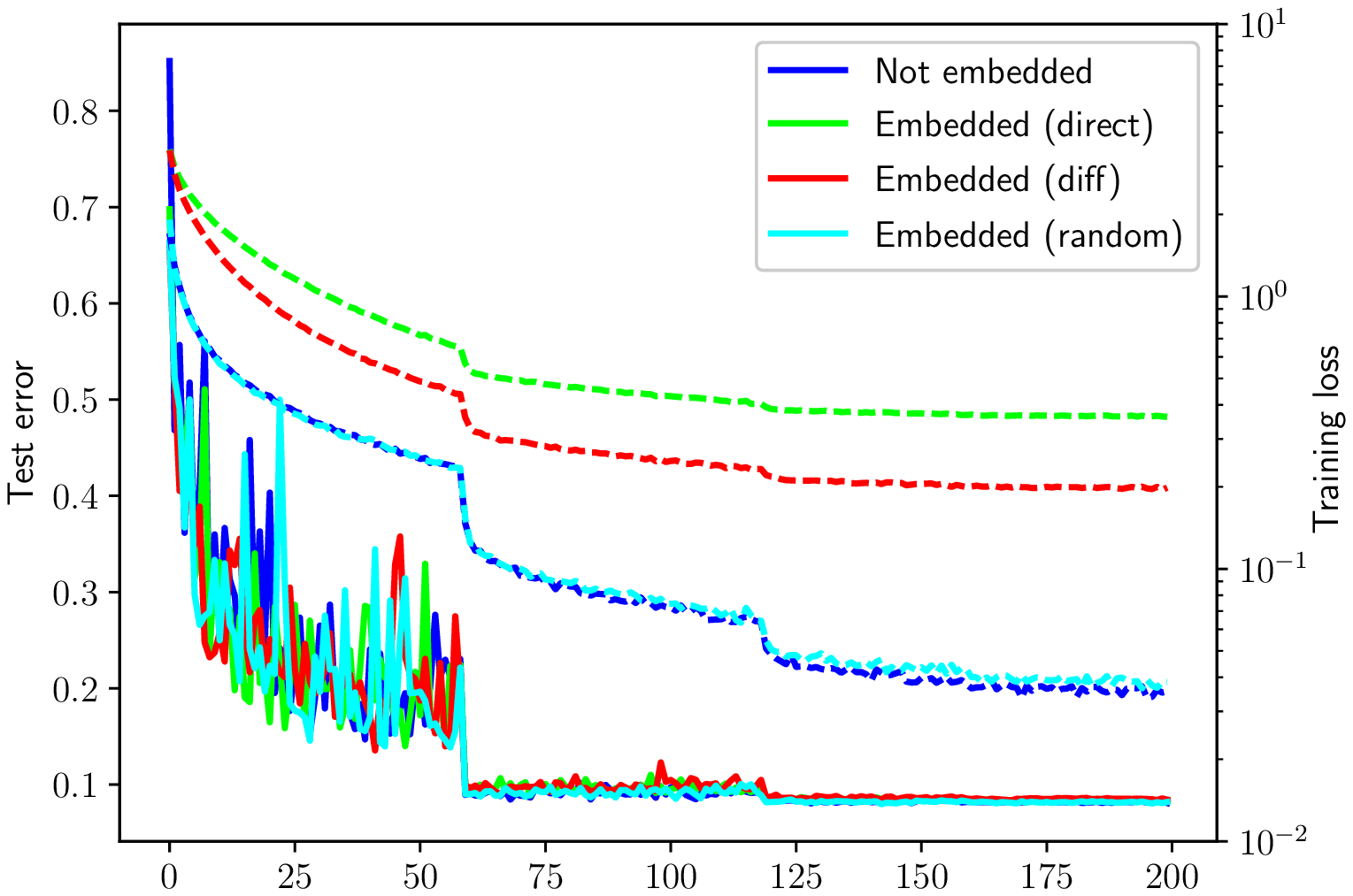}
	\caption{Training curves for the host network on CIFAR-10 as a function of epochs.
	Solid lines denote test error (y-axis on the left) and dashed lines denote training loss $E(\bm{w})$ (y-axis on the right).}
	\label{fig:history}
\end{figure}

Figure~\ref{fig:history} shows the training curves for the host network in CIFAR-10 as a function of epochs.
\textsf{Not embedded} is the case where the host network is trained without the embedding regularizer.
\textsf{Embedded (direct)}, \textsf{Embedded (diff)}, and \textsf{Embedded (random)} respectively represent training curves with embedding regularizers whose parameters are $\bm{X}^{\textsf{direct}}$, $\bm{X}^{\textsf{diff}}$, and $\bm{X}^{\textsf{random}}$.
We can see that the training loss $E(\bm{w})$ with a watermark becomes larger than the not-embedded case if the parameters $\bm{X}^{\textsf{direct}}$ and $\bm{X}^{\textsf{diff}}$ are used.
This large training loss is dominated by the embedding loss $E_R (\bm{w})$, which indicates that it is difficult to embed a watermark directly into a parameter or even into the difference of two parameters.
On the other hand, the training loss of \textsf{Embedded (random)} is very close to that of \textsf{Not embedded}.

Table~\ref{tab:embed_result} shows the best test errors and embedding losses $E_R (\bm{w})$ of the host networks with and without embedding.
We can see that the test errors of \textsf{Not embedded} and \textsf{random} are almost the same while those of \textsf{direct} and \textsf{diff} are slightly larger.
The embedding loss $E_R (\bm{w})$ of \textsf{random} is extremely low compared with those of \textsf{direct} and \textsf{diff}.
These results indicate that the \textsf{random} approach can effectively embed a watermark without impairing the performance in the original task.

\begin{comment}
\begin{figure}[tb]
	\centering
	\includegraphics[width=\linewidth]{history_finetune.eps}
	\caption{The training loss and test error as a function of epochs.}
	\label{fig:history}
\end{figure}
\end{comment}

\begin{table}[tb]
	\centering
	\caption{Test error ($\%$) and embedding loss $E_R (\bm{w})$ with and without embedding.}
	\label{tab:embed_result}
\begin{tabular}{c|cc} \hline
				& Test error	& $E_R (\bm{w})$	\\ \hline
Not embedded	& 8.04				& N/A				\\
direct	& 8.21			& $1.24{\times}10^{-1}$	\\
diff	& 8.37			& $6.20{\times}10^{-2}$	\\
random	& 7.97			& $4.76{\times}10^{-4}$	\\ \hline
	\end{tabular} \\
\end{table}

\subsubsection{Fine-tune-to-embed and Distill-to-embed}
\label{sec:res_retrain}
In the above experiments, a watermark was embedded by training the host network from scratch (train-to-embed).
Here, we evaluated the other two situations introduced in Section~\ref{sec:situation}: fine-tune-to-embed and distill-to-embed.

For fine-tune-to-embed, two experiments were performed.
In the first experiment, the host network was trained on the CIFAR-10 dataset without embedding, and then fine-tuned on the same CIFAR-10 dataset with and without embedding (for comparison).
In the second experiment, the host network is trained on the Caltech-101 dataset, and then fine-tuned on the CIFAR-10 dataset with and without embedding.

\begin{figure}[tb]
	\centering
	\includegraphics[width=\linewidth]{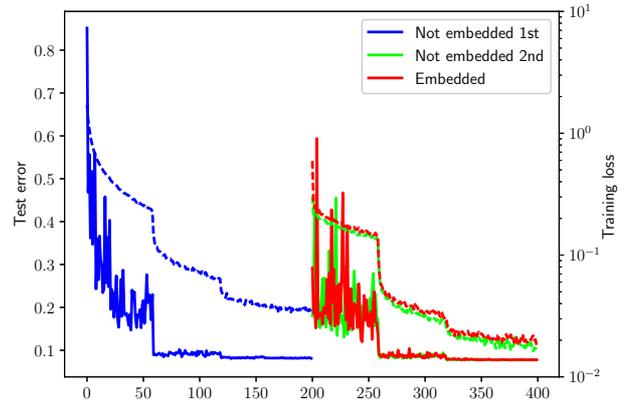}
	\caption{Training curves for fine-tuning the host network.
	The first and second halves of epochs correspond to the first and second training.
	Solid lines denote test error (y-axis on the left) and dashed lines denote training loss (y-axis on the right).}
	\label{fig:history_finetune_embed_same}
\end{figure}

Table~\ref{tab:embed_result_finetune} (a) shows the result of the first experiment.
\textsf{Not embedded 1st} corresponds to the first training without embedding.
\textsf{Not embedded 2nd} corresponds to the second training without embedding and \textsf{Embedded} corresponds to the second training with embedding.
Figure~\ref{fig:history_finetune_embed_same} shows the training curves of these fine-tunings\footnote{Note that the learning rate was also initialized to 0.1 at the beginning of the second training, while the learning rate was reduced to $8.0 \times 10^{-4}$) at the end of the first training.}.
We can see that \textsf{Embedded} achieved almost the same test error as \textsf{Not embedded 2nd} and a very low $E_R (\bm{w})$.

\begin{figure}[tb]
	\centering
	\includegraphics[width=\linewidth]{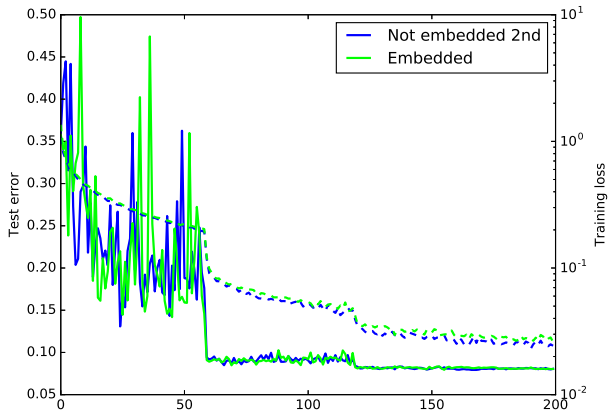}
	\caption{Training curves for the host network on CIFAR-10 as a function of epochs.
	Solid lines denote test error (y-axis on the left) and dashed lines denote training loss (y-axis on the right).}
	\label{fig:history_finetune_embed_different}
\end{figure}

Table~\ref{tab:embed_result_finetune} (b) shows the results of the second experiment.
\textsf{Not embedded 2nd} corresponds to the second training without embedding and \textsf{Embedded} corresponds to the second training with embedding.
Figure~\ref{fig:history_finetune_embed_different} shows the training curves of these fine-tunings.
The test error and training loss of the first training are not shown because they are not compatible with the two different training datasets.
From these results, it was also confirmed that \textsf{Embedded} achieved almost the same test error as \textsf{Not embedded 2nd} and very low $E_R (\bm{w})$.
Thus, we can say that the proposed method is effective even in the fine-tune-to-embed situation (in the same and different domains).

Finally, embedding a watermark in the distill-to-embed situation was evaluated.
The host network is first trained on the CIFAR-10 dataset without embedding.
Then, the trained network was further fine-tuned on the same CIFAR-10 dataset with and without embedding.
In this second training, the training labels of the CIFAR-10 dataset were \textit{not} used.
Instead, the predicted values of the trained network were used as soft targets~\cite{hin_nipsw14}.
In other words, no label was used in the second training.
Table~\ref{tab:embed_result_finetune} (c) shows the results for the distill-to-embed situation.
\textsf{Not embedded 1st} corresponds to the first training and \textsf{Embedded} (\textsf{Not embedded 2nd}) corresponds to the second distilling training with embedding (without embedding).
It was found that the proposed method also achieved low test error and $E_R (\bm{w})$ in the distill-to-embed situation.
Table~\ref{tab:embed_result_finetune} (d) shows the result for the distill-to-embed situation on the different domain; the difference from Table~\ref{tab:embed_result_finetune} (c) is that the predicted values for the Caltech-101 are used as soft targets here instead of CIFAR-10.
The test error is calculated on CIFAR-10.

\begin{table}[tb]
	\centering
	\caption{Test error ($\%$) and embedding loss $E_R (\bm{w})$ with and without embedding in fine-tuning and distilling.}
	\label{tab:embed_result_finetune}
	(a) Fine-tune-to-embed (CIFAR-10 $\rightarrow$ CIFAR-10) \\ 
\begin{tabular}{c|cc} \hline
					& Test error	& $E_R (\bm{w})$	\\ \hline
Not embedded 1st	& 8.04			& N/A				\\
Not embedded 2nd	& 7.66			& N/A				\\
Embedded			& 7.70			& $4.93{\times}10^{-4}$	\\ \hline
	\end{tabular} \\  %\vspace{3mm}
		(b) Fine-tune-to-embed (Caltech-101 $\rightarrow$ CIFAR-10) \\ 
\begin{tabular}{c|cc} \hline
					& Test error	& $E_R (\bm{w})$	\\ \hline
Not embedded 2nd		& 7.93			& N/A				\\
Embedded			& 7.94			& $4.83{\times}10^{-4}$	\\ \hline
	\end{tabular} \\  % \vspace{3mm}
	(c) Distill-to-embed (CIFAR-10 $\rightarrow$ CIFAR-10)	\\ 
\begin{tabular}{c|cc} \hline
				& Test error	& $E_R (\bm{w})$	\\ \hline
Not embedded 1st	& 8.04				& N/A				\\
Not embedded 2nd	& 7.86				& N/A				\\
Embedded			& 7.75			& $5.01{\times}10^{-4}$	\\ \hline
	\end{tabular} \\  % \vspace{3mm}
	(d) Distill-to-embed (CIFAR-10 $\rightarrow$ Caltech-101)	\\ 
	\begin{tabular}{c|cc} \hline
				& Test error	& $E_R (\bm{w})$	\\ \hline
Not embedded 1st	& 8.04				& N/A				\\
Embedded			& 28.34			& $5.80{\times}10^{-3}$	\\ \hline
	\end{tabular} \\
\end{table}

\subsection{Robustness of Embedded Watermarks}
In this section, the robustness of the proposed watermark is evaluated for the three types of attacks explained in Section~\ref{sec:attack}: fine-tuning, model compression, and watermark overwriting.

\subsubsection{Robustness against Fine-tuning}
\label{sec:finetune}
Fine-tuning or transfer learning~\cite{Simonyan_iclr15} seems to be the most likely type of (unintentional) attack because it is frequently performed on trained models to apply them to other but similar tasks with less effort than training a network from scratch or to avoid over-fitting when sufficient training data are not available.

In this experiment, two trainings were performed; in the first training, a 256-bit watermark was embedded in the \textsf{conv 2} group in the train-to-embed manner, and then the host network was further fine-tuned in the second training without embedding, to determine whether or not the watermark embedded in the first training stayed in the host network, even after the second training (fine-tuning).

Table~\ref{tab:robust_finetune} shows the embedding loss before fine-tuning ($E_R (\bm{w})$) and after fine-tuning ($E'_R (\bm{w})$), and the best test error after fine-tuning.
In the same domain, the host network is trained on the CIFAR-10 dataset while embedding a watermark, and then further fine-tuned without embedding a watermark.
We evaluated fine-tuning in the same domain (CIFAR-10 $\rightarrow$ CIFAR-10) and in the different domains (Caltech-101 $\rightarrow$ CIFAR-10).
We can see that, in both cases, the embedding loss was increased slightly by fine-tuning but was still low.
In addition, the bit error rate of the detected watermark was equal to zero in both cases.
The reason why the embedding loss in fine-tuning in the different domains is higher than that in the same domain is that the Caltech-101 dataset is significantly more difficult than the CIFAR-10 dataset in our settings; all images in the Caltech-101 dataset were resized to $32 \times 32$\footnote{This size is extremely small compared with their original sizes (roughly $300 \times 200$).} for compatibility with the CIFAR-10 dataset.

\begin{comment}
\textbf{the same training dataset} \textbf{a different training dataset from the same domain}, and \textbf{a training dataset from the different domain}. 
\begin{enumerate}
\item the same training dataset
\item a different training dataset from the same domain
\item a training dataset from the different domain
\end{enumerate}
\end{comment}

\begin{table}[tb]
	\centering
	\caption{Embedding loss before fine-tuning ($E_R (\bm{w})$) and after fine-tuning ($E'_R (\bm{w})$), and the best test error ($\%$) and bit error rate (BER) after fine-tuning.}
	\label{tab:robust_finetune}
\begin{tabular}{c|c@{}c@{}c@{}c@{}} \hline
					& $E_R (\bm{w})$				& $E'_R (\bm{w})$			&BER& Test error	\\ \hline
CIFAR-10 $\rightarrow$ CIFAR-10	& $4.76{\times}10^{-4}$	& $8.66{\times}10^{-4}$	&0.00& 7.69	\\
Caltech-101 $\rightarrow$ CIFAR-10	& $5.96{\times}10^{-3}$	& $1.56{\times}10^{-2}$	&0.00& 7.88	\\ \hline
	\end{tabular} \\
\end{table}

\subsubsection{Robustness against Model Compression}
It is sometimes difficult to deploy deep neural networks in embedded systems or mobile devices because they are both computationally intensive and memory intensive.
In order to solve this problem, the model parameters are often \textit{compressed}~\cite{han_nips15, han_isca16, han_iclr16}.
The compression of model parameters can intentionally or unintentionally act as an attack against watermarks.
In this section, we evaluate the robustness of our watermarks against model compression, in particular, against parameter pruning~\cite{han_nips15} and distillation~\cite{hin_nipsw14}.

\textbf{Robustness against parameter pruning.}
In parameter pruning, parameters whose absolute values are very small are cut-off to zero.
In~\cite{han_iclr16}, quantization of weights and the Huffman coding of quantized values are further applied.
Because quantization has less impact than parameter pruning and the Huffman coding is lossless compression, we focus on parameter pruning.

\begin{figure}[tb]
	\centering 
	\begin{minipage}[b]{\linewidth}
		\includegraphics[width=\linewidth]{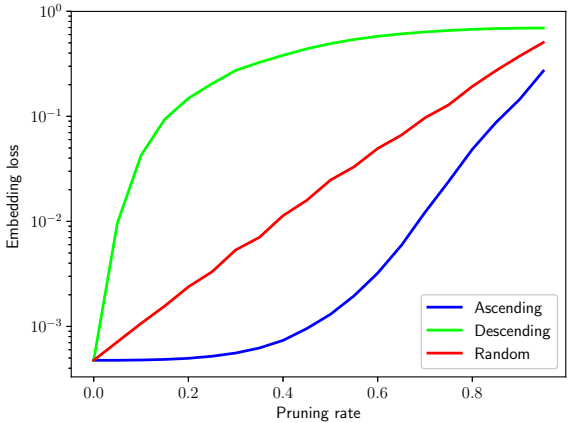} \\
	\centering (a) Embedding loss.
	\end{minipage}
	\begin{minipage}[b]{\linewidth}
		\includegraphics[width=\linewidth]{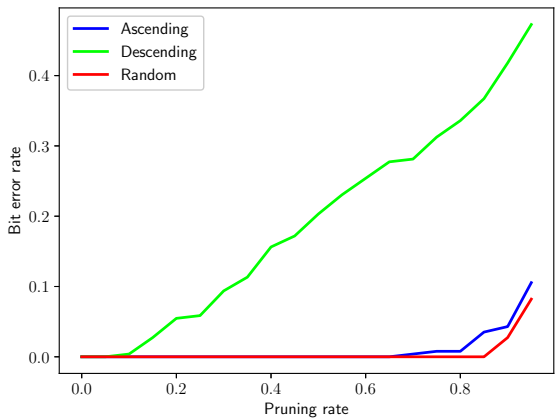} \\
	\centering (b) Bit error rate.
	\end{minipage}
	\caption{Embedding loss and bit error rate after pruning as a function of pruning rate.}
	\label{fig:prune}
\end{figure}

In order to evaluate robustness against parameter pruning, we embedded a 256-bit watermark in the \textsf{conv 2} group while training the host network on the CIFAR-10 dataset.
We removed $\alpha$\% of the $3 \times 3 \times 64 \times 64$ parameters of the embedded layer and calculated embedding loss and bit error rate.
Figure~\ref{fig:prune} (a) shows embedding loss $E_R (\bm{w})$ as a function of pruning rate $\alpha$.
\textsf{Ascending} (\textsf{Descending}) represents embedding loss when the top $\alpha$\% parameters are cut-off according to their absolute values in ascending (descending) order.
\textsf{Random} represents embedding loss where $\alpha$\% of parameters are randomly removed.
\textsf{Ascending} corresponds to parameter pruning and the others were evaluated for comparison.
We can see that the embedding loss of \textsf{Ascending} increases more slowly than those of \textsf{Descending} and \textsf{Random} as $\alpha$ increases.
It is reasonable that model parameters with small absolute values have less impact on a detected watermark because the watermark is extracted from the dot product of the model parameter $w$ and the constant embedding parameter (weight) $\bm{X}$.

\begin{figure}[tb]
	\centering
	\includegraphics[width=\linewidth]{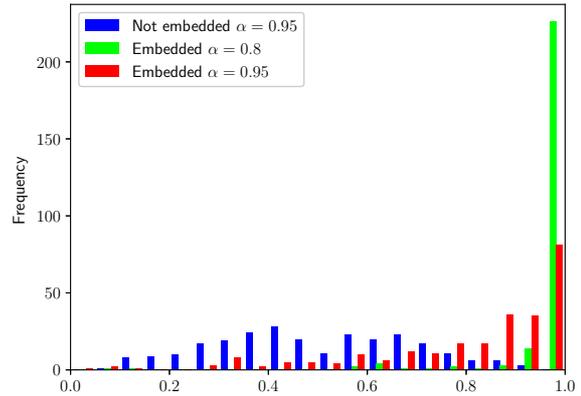}
	\caption{Histogram of the detected watermark $\sigma(\sum_{i} X_{ji} w_i)$ after pruning.}
	\label{fig:hist_pruned}
\end{figure}

Figure~\ref{fig:prune} (b) shows the bit error rate as a function of pruning rate $\alpha$.
Surprisingly, the bit error rate was still zero after removing 65\% of the parameters and $2/256$ even after 80\% of the parameters were pruned (\textsf{Ascending}).
We can say that the embedded watermark is sufficiently robust against parameter pruning because, in~\cite{han_iclr16}, the resulting pruning rate of convolutional layers ranged from to 16\% to 65\% for the AlexNet~\cite{kri_nips12}, and from 42\% to 78\% for VGGNet~\cite{Simonyan_iclr15}.
Furthermore, this degree of bit error can be easily corrected by an error correction code (e.g. the BCH code).
Figure~\ref{fig:hist_pruned} shows the histogram of the detected watermark $\sigma(\sum_{i} X_{ji} w_i)$ after pruning for $\alpha = 0.8$ and $0.95$.
For $\alpha = 0.95$, the histogram of the detected watermark is also shown for the host network into which no watermark is embedded.
We can see that many of $\sigma(\sum_{i} X_{ji} w_i)$ are still close to one for the embedded case, which might be used as a confidence score in determining the existence of a watermark (zero-bit watermarking).

\textbf{Robustness against distillation.}
Distillation is a training procedure initially designed to train a deep neural networks model using knowledge transferred from a different model.
The intuition was suggested in~\cite{dimmy_nips14} while distillation itself was formally introduced in~\cite{hin_nipsw14}. 
Distillation is employed to reduce computational complexity or compressing the knowledge in an ensemble of models into a single small model. 
In the standard distillation framework, a large network (or multiple networks) is first trained and then a smaller network is trained using the predicted labels of the large network in order to compress the large network. 
As well as fine-tuning, distillation could be an unintentional attack and it is specific to deep neural networks. 

In this experiment, we performed two trainings. First a 256-bit watermark was embedded in the \textsf{conv 2} group in the train-to-embed manner with CIFAR-10. Then, in the second training, another model was distilled using the CIFAR-10 dataset and the predicted values of the first trained network instead of the actual labels. The second training did not embed a watermark and initial weights were set at random. 
We employed the simplest form of distillation in this experiment. Although we could use a different network architecture and different dataset in the transfer step, we trained a new model of the same architecture on the same set CIFAR-10 for simplicity. 

Table~\ref{tab:distill_atack} shows the test error and bit error rate after the first and second training. 
The watermark could not be detected from the distilled model as expected because the model weights had been initialized with random weights.

\begin{table}[tb]
	\centering
	\caption{Test error ($\%$), and bit error rate (BER)  of the embedded host network and after distilling without embedding the watermark.}
	\label{tab:distill_atack}
\begin{tabular}{c|cc} \hline
							&	Test error		&	BER	\\ \hline
Embedded 1st					& 	8.05			& 	0.00	\\
After distillation		& 	8.40			& 	0.54	\\ \hline
	\end{tabular} \\
\end{table}

\subsubsection{Robustness against Watermark Overwriting}
\label{sec:overwrite}
Overwriting is a common attack in digital content watermarking~\cite{neil_01}.
A third-party user may embed a different watermark in order to overwrite the original watermark.
Basically, it is necessary to know where the original watermark is embedded to overwrite watermarks.
Please note that in addition to regularizer parameters $\bm{X}$, which work as a secret key, the location where a digital watermark is embedded should be also be secret information.
However, it is conceivable for a watermark  to be embedded into all or multiple layers to destroy the embedded original watermark or change ownership without exact information on where the original watermark is actually embedded. 

In order to evaluate robustness against overwriting, we embedded a 256-bit watermark in the \textsf{conv 2},  \textsf{conv 3} and \textsf{conv 4} groups with a regularizer parameter $\bm{X}_0$, while training the host network on the CIFAR-10 dataset.  Then, we additionally embedded a 256-bit, 512-bit, 1024-bit and 2048-bit watermark into the host network respectively with a regularizer parameter $\bm{X}_0$  different from $\bm{X}_1$. 
The number of parameters $\bm{w}$ of \textsf{conv 2}, \textsf{conv 3}, and \textsf{conv 4} groups were 576, 1152,  and 2304, respectively. All bit error rates of the original host networks were zero. The additional watermarks were embedded while training on the CIFAR-10 dataset.  

Table~\ref{tab:overwrite} shows test error, embedding loss $E_R (\bm{w})$ and bit error rate with the first regularizer parameter $\bm{X}_0$ after overwriting the first watermark. 
When the bit error rate is close to 0.5, it indicates that the original watermark has been erased completely.
We can see that the original watermark was erased in some cases where the number of embedded bits was large compared to the number of parameters $\bm{w}$. 

\begin{table}[tb]  
\centering
\caption{Test error ($\%$), embedding loss $E_R (\bm{w})$ and bit error rate with the original regularizer parameter after overwriting a watermark. The number of parameters $\bm{w}$ of \textsf{conv 2}, \textsf{conv 3}, and \textsf{conv 4} groups are 576, 1152,  and 2304, respectively.}
\label{tab:overwrite}
(a) Test error ($\%$) \\
\begin{tabular}{c|cccc}  \hline
Embedded bits	& \multicolumn{3}{c}{Embedded group}	\\
& conv 2	& conv 3	& conv 4	\\ \hline
256				& $7.43$	& $7.36$	& $7.96$	\\
512				& $7.29$	& $7.35$	& $7.92$	\\
1,024			& $7.58$	& $7.41$	& $7.96$	\\
2,048			& $7.36$	& $7.61$	& $7.94$	\\ \hline
\end{tabular} \\  % \vspace{3mm}
(b) Embedding loss \\
\begin{tabular}{c|llll} \hline
Embedded bits	& \multicolumn{3}{c}{Embedded group}	\\
& conv 2	& conv 3	& conv 4	\\ \hline

256				& 1.67		& $2.05{\times}10^{-1}$		& $4.98{\times}10^{-2}$	\\
512				& 4.28		& 1.13		& $1.94{\times}10^{-1}$	\\
1,024			& $1.77{\times}10^{1}$		& 3.76		& $5.24{\times}10^{-1}$	\\
2,048			& 1.04		& $1.12{\times}10^{1}$		& 1.40	\\ \hline
\end{tabular} \\ % \vspace{3mm}
(c) Bit error rate \\  
\begin{tabular}{c|cccc} \hline
Embedded bits	& \multicolumn{3}{c}{Embedded group}	\\
& conv 2	& conv 3	& conv 4	\\ \hline
256				& $3.09{\times}10^{-1}$	& $8.59{\times}10^{-2}$	& $3.90{\times}10^{-3}$	\\
512				& $4.10{\times}10^{-1}$	& $2.38{\times}10^{-1}$	& $6.64{\times}10^{-2}$	\\
1,024			& $5.11{\times}10^{-1}$	& $4.29{\times}10^{-1}$	& $1.99{\times}10^{-1}$	\\
2,048			& $5.27{\times}10^{-1}$	& $5.07{\times}10^{-1}$	& $3.55{\times}10^{-1}$	\\ \hline
\end{tabular} \\
\end{table}

\subsection{Capacity of Watermark.}
\label{sec:capacity}
In this section, the capacity of the embedded watermark is explored by embedding different sizes of watermarks into different groups in the train-to-embed manner.
Please note that the number of parameters $\bm{w}$ of \textsf{conv 2}, \textsf{conv 3}, and \textsf{conv 4} groups were 576, 1152,  and 2304, respectively.
Table~\ref{tab:capacity} shows test error ($\%$), embedding loss $E_R (\bm{w})$ and bit error rate for combinations of different embedded blocks and different numbers of embedded bits.
We can see that embedded loss or test error becomes high if the number of embedded bits becomes larger than the number of parameters $\bm{w}$ (e.g. 2,048 bits in \textsf{conv 3}) because the embedding problem becomes overdetermined in such cases.
Thus, the number of embedded bits should be smaller than the number of parameters $\bm{w}$, which is a limitation of the embedding method using a single-layer perceptron.
This limitation would be resolved by using a multi-layer perceptron in the embedding regularizer.

\begin{table}[tb]
	\centering
	\caption{Test error ($\%$), embedding loss $E_R (\bm{w})$ and bit error rate for the combinations of embedded groups and sizes of embedded bits. The number of parameters $\bm{w}$ of \textsf{conv 2}, \textsf{conv 3}, and \textsf{conv 4} groups were 576, 1152,  and 2304, respectively.}
	\label{tab:capacity}
	(a) Test error ($\%$) \\
	\begin{tabular}{c|cccc} \hline
Embedded bits	& \multicolumn{3}{c}{Embedded group}	\\
& conv 2	& conv 3	& conv 4	\\ \hline
256				& 7.97		& 7.98		& 7.92	\\
512				& 8.47		& 8.22		& 7.84	\\
1,024			& 8.43		& 8.12		& 7.84	\\
2,048			& 8.17		& 8.93		& 7.75	\\ \hline
	\end{tabular} \\  % \vspace{3mm}
(b) Embedding loss \\ 
	\begin{tabular}{c|cccc} \hline
Embedded bits	& \multicolumn{3}{c}{Embedded group}	\\
& conv 2	& conv 3	& conv 4	\\ \hline
256				& $4.76{\times}10^{-4}$	& $7.20{\times}10^{-4}$	& $1.10{\times}10^{-2}$	\\
512				& $8.11{\times}10^{-4}$	& $8.18{\times}10^{-4}$	& $1.25{\times}10^{-2}$	\\
1,024			& $6.74{\times}10^{-2}$	& $1.53{\times}10^{-3}$	& $1.53{\times}10^{-2}$	\\
2,048			& $5.35{\times}10^{-1}$	& $3.70{\times}10^{-2}$	& $3.06{\times}10^{-2}$	\\ \hline
	\end{tabular} \\  % \vspace{3mm}
(c) Bit error rate \\ 
	\begin{tabular}{c|cccc} \hline
Embedded bits	& \multicolumn{3}{c}{Embedded group}	\\
& conv 2	& conv 3	& conv 4	\\ \hline
256				& 	0.00	&	0.00 	& 0.00		\\
512				&	0.00 	&	0.00	& 0.00		\\
1,024			&	0.00	&	0.00 	& 0.00		\\
2,048			&	0.28 	&	0.00 	& 0.00		\\ \hline
	\end{tabular} \\
\end{table}

\section{Discussion}
\subsection{Insights}
　
\indent\textbf{Fidelity.}
As mentioned in Section~\ref{sec:regularizer}, poor local minima are rarely a problem with large networks in practice. Regardless of the initial conditions, the system nearly always reaches solutions of very similar quality. Recent theoretical and empirical results strongly suggest that local minima are not a serious issue in general~\cite{lecun_nature15}. Therefore, the proposed approach was able to maintain the performance of the original task and carry out successful watermarking as shown in the experimental results of Section~\ref{sec:res_train} and Section~\ref{sec:res_retrain}.  

\textbf{Robustness.}
For watermarking techniques in the neural networks domain, fine-tuning seems to be the most feasible and significant attack. 
The experimental results in Section~\ref{sec:finetune} show the proposed method could retain the watermark completely after fine-tuning in both cases: the same domain and a different domain.
In the case of the same domain, updates of weight values were assumed to be small if the host model was trained well in the first training. 
On the other hand, in the case of a different domain, weight values are supposed to change dramatically.
However, our experimental results show the watermark remained after fine-tuning to a different domain.
It is considered that fine-tuning would cause less alteration for weights near the input layer compared to near the output layer.
Therefore, the digital watermark could successfully resist a fine-tuning attack, if the watermark is embedded near the input layer of sufficiently deep networks.
Additionally,  there is an advantage that the network configuration near the input layer may not be changed for another task.

\indent\textbf{Capacity.}
The result presented in Section~\ref{sec:capacity} indicates that the capacity is strongly related to the number of the host weights compared to the length of watermarks. Capacity may be increased by using a multi-layer perceptron in the embedding regularizer.

\subsection{Limitations}
Although we have obtained some initial insight into the new problem of embedding a watermark in deep neural networks, 
the proposed approach still has the following limitations.

\indent\textbf{Distillation.} 
Distillation is theoretically a serious attack for watermarking of neural networks. 
However, distillation does not seem to be an important attack in reality, since it requires data that are very similar to the inputs used in the original training phase in order to maintain fidelity.

\textbf{Overwriting.} 
As shown in Section~\ref{sec:overwrite}, overwriting destroys the original watermark. This experiment is assumed to know exactly where the original watermark was embedded. It is conceivable that watermarks could be embedded into all or multiple layers to destroy the original watermark, although this would incur a much greater computational cost due to the large size of widely targeted parameters. 
Overwriting is still a serve attack  and we should explore an effective way of combatting overwriting.

\textbf{Black-box type situation.} 
In the proposed digital watermarking approach for deep neural network models, we make an assumption that the weight values are visible. 
Thus, it is impossible to detect abuse in a black-box type situation such as a client-server system where a watermarked model is used on a server by unauthorized parties. To effectively deal with such a situation, the copyright protection of neural network models requires another approach. 
Inspired by our work~\cite{uchida_icmr17}, Merrer et al. propose a method that allows the extraction of the watermark from a neural network remotely through a service API~\cite{erwan_arxiv17}. The method embeds zero-bit watermarks into models with a stitching algorithm based on adversaries. 

\subsection{Further Expected Developments}
Further developments are expected by using the analogy of digital content protection and domain-specific issues for deep neural networks.

\textbf{Embedding as sequential learning.}
In Section~\ref{sec:direct}, we have shown that it is not effective to directly embed a watermark without considering the original task. We can consider this embedding process as sequential learning; the training of the original task is the first task, and subsequent watermark embedding is the second task. Thus, the increase of error rate after embedding can be interpreted as \textit{catastrophic forgetting}~\cite{kirkpatrick_17}. From this point of view, we can adopt recently developed methods~\cite{lee_nips17, kirkpatrick_17} to overcome this catastrophic forgetting in embedding watermark.

\indent\textbf{Compression as embedding.}
Compressing deep neural networks is a very important and active research topic.
While we confirmed in this paper that our watermark is very robust against parameter pruning in this paper, a watermark might be embedded in conjunction with compressing models.
For example, in~\cite{han_iclr16}, after parameter pruning, the network is re-trained to learn the final weights for the remaining sparse parameters.
Our embedding regularizer can be used in this re-training to embed a watermark.

\textbf{Network morphism.}
In~\cite{Chen_iclr16, Wei_icml16}, a systematic study has been conducted on how to morph a well-trained neural network into a new one so that its network function can be completely preserved for further training.
This network morphism can constitute a severe attack against our watermark because it may be impossible to detect the embedded watermark if the topology of the host network undergoes major modification.
We have left the investigation into how the embedded watermark is affected by this network morphism as a topic for future work.

\textbf{Steganalysis.} Steganalysis~\cite{shaohui_icme03, kodovsky_tifs12} is a method for detecting the presence of secretly hidden data (e.g. steganography or watermarks) in digital media files such as images, video, audio, and, in our case, deep neural networks.
Watermarks ideally are robust against steganalysis.
While, in this paper, we confirmed that embedding watermarks does not significantly change the distribution of model parameters, more exploration is needed to evaluate robustness against steganalysis.
Conversely, developing effective steganalysis against watermarks for deep neural networks could be an interesting research topic.

\textbf{Fingerprinting.} Digital fingerprinting is an alternative to the watermarking approach for persistent identification of images~\cite{bar_icassp03}, video~\cite{jol05, uch_icmr11}, and audio clips~\cite{ang_icme12, hai02}.
In this paper, we focused on one of these two important approaches.
Robust fingerprinting of deep neural networks is another and complementary direction to protect deep neural network models.

\section{Conclusions}
In this paper, we have proposed a general framework for embedding a watermark in deep neural network models to protect the rights to the trained models.
First, we formulated a new problem: embedding watermarks into deep neural networks.
We also defined requirements, embedding situations, and the types of attacks that watermarking deep neural networks are vulnerable to.
Second, we proposed a general framework for embedding a watermark in model parameters using a parameter regularizer.
Our approach does not impair the performance of networks into which a watermark is embedded.
Finally, we performed comprehensive experiments to reveal the potential of watermarking deep neural networks as the basis of this new problem.
We showed that our framework could embed a watermark without impairing the performance of a deep neural network.
The embedded watermark did not disappear even after fine-tuning or parameter pruning; the entire watermark remained even after 65\% of the parameters were pruned.

\balance

%\footnotesize
%\bibliographystyle{IEEEtran}
\bibliographystyle{abbrv}
\bibliography{refs}

%\begin{acknowledgements}
%\end{acknowledgements}


\end{document}